\documentclass{article}

     \PassOptionsToPackage{numbers, compress}{natbib}
    \usepackage[final]{neurips_2020}

\usepackage[utf8]{inputenc} %
\usepackage[T1]{fontenc}    %
\usepackage{hyperref}       %
\usepackage{url}            %
\usepackage{booktabs}    %
\usepackage{amsfonts}       %
\usepackage{nicefrac}       %
\usepackage{microtype}      %
\usepackage{todonotes}
\usepackage{subcaption}
\usepackage{xspace}
\usepackage{amsmath}
\usepackage{amssymb}
\usepackage{bbm}
\usepackage{multirow}
\newtheorem{theorem}{Theorem}

\newtheorem{proof}{Proof}[section]

\usepackage{floatrow}
\newfloatcommand{capbtabbox}{table}[][\FBwidth]

\usepackage{float}

\newfloat{sbsfigure}{t}{ext}

\DeclareMathOperator*{\argmax}{arg\,max}

\newcommand{\ie}{i.e.,\xspace}
\newcommand{\eg}{e.g.,\xspace}
\newcommand{\eat}[1]{}

\newcommand{\baby}{Patience-based Early Exit\xspace}
\newcommand{\abbr}{PABEE\xspace}

\title{BERT Loses Patience:\\ Fast and Robust Inference with Early Exit}

\author{%
  Wangchunshu Zhou$^1$\thanks{Equal contribution. Work done during these two authors' internship at Microsoft Research Asia.}, Canwen Xu$^{2*}$, Tao Ge$^3$, Julian McAuley$^2$, Ke Xu$^1$, Furu Wei$^3$ \\
  $^1$Beihang University $^2$University of California, San Diego $^3$Microsoft Research Asia \\
  $^1$\texttt{zhouwangchunshu@buaa.edu.cn,kexu@nlsde.buaa.edu.cn} \\
  $^2$\texttt{\{cxu,jmcauley\}@ucsd.edu} $^3$\texttt{\{tage,fuwei\}@microsoft.com} \\
}

\begin{document}

\maketitle

\begin{abstract}

In this paper, we propose \baby , a straightforward yet effective inference method that can be used as a plug-and-play technique to simultaneously improve the efficiency and robustness of a pretrained language model (PLM). To achieve this, our approach couples an internal-classifier with each layer of a PLM and dynamically stops inference when the intermediate predictions of the internal classifiers remain unchanged for a pre-defined number of steps.
Our approach improves inference efficiency as it allows the model to make a prediction with fewer layers. Meanwhile, experimental results with an ALBERT model show that our method can improve the accuracy and robustness of the model by preventing it from overthinking and exploiting multiple classifiers for prediction, yielding a better accuracy-speed trade-off compared to existing early exit methods.\footnote{Code available at \url{https://github.com/JetRunner/PABEE}.}
\end{abstract}

\section{Introduction}
In Natural Language Processing (NLP), pretraining and fine-tuning have become a new norm for many tasks. Pretrained language models (PLMs) (\eg BERT~\cite{devlin2018bert}, XLNet~\cite{yang2019xlnet}, RoBERTa~\cite{liu2019roberta}, ALBERT~\cite{lan2019albert}) contain many layers and millions or even billions of parameters, making them computationally expensive and inefficient regarding both memory consumption and latency. This drawback hinders their application in scenarios where inference speed and computational costs are crucial. Another bottleneck of overparameterized PLMs that stack dozens of Transformer layers is the ``overthinking'' problem~\cite{kaya2018shallow} during their decision-making process. That is, for many input samples, their shallow representations at an earlier layer are adequate to make a correct classification, whereas the representations in the final layer may be otherwise distracted by over-complicated or irrelevant features that do not generalize well. The overthinking problem in PLMs leads to wasted computation, hinders model generalization, and may also make them vulnerable to adversarial attacks~\cite{jin2019bert}.

\begin{figure}
  \centering
  \begin{subfigure}[t]{0.46\textwidth}
        \centering
        \includegraphics[width=\textwidth]{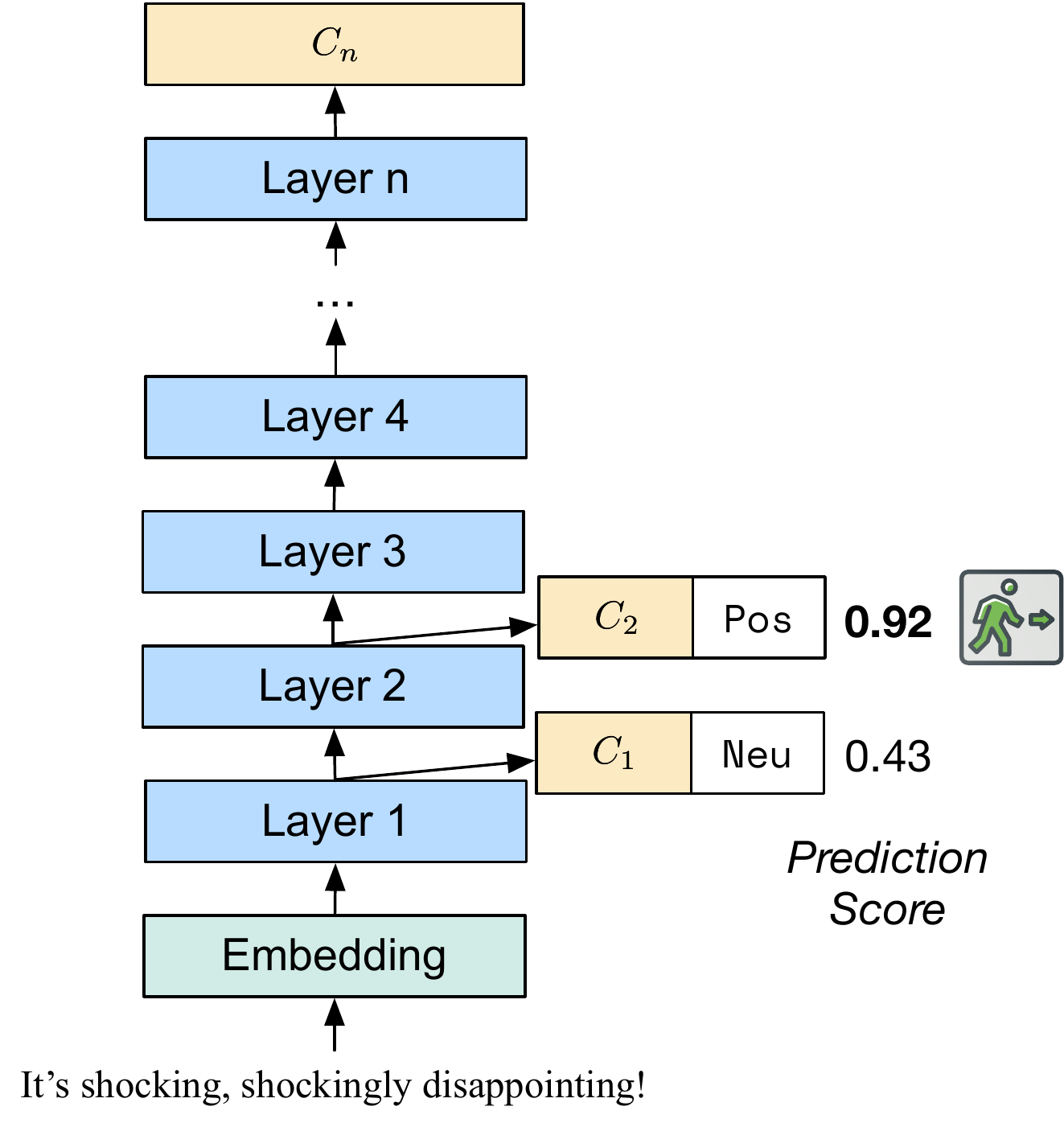}
        \caption{\label{fig:sdn}Shallow-Deep Net~\cite{kaya2018shallow}}
    \end{subfigure}%
    ~
    \begin{subfigure}[t]{0.46\textwidth}
        \centering
        \includegraphics[width=\textwidth]{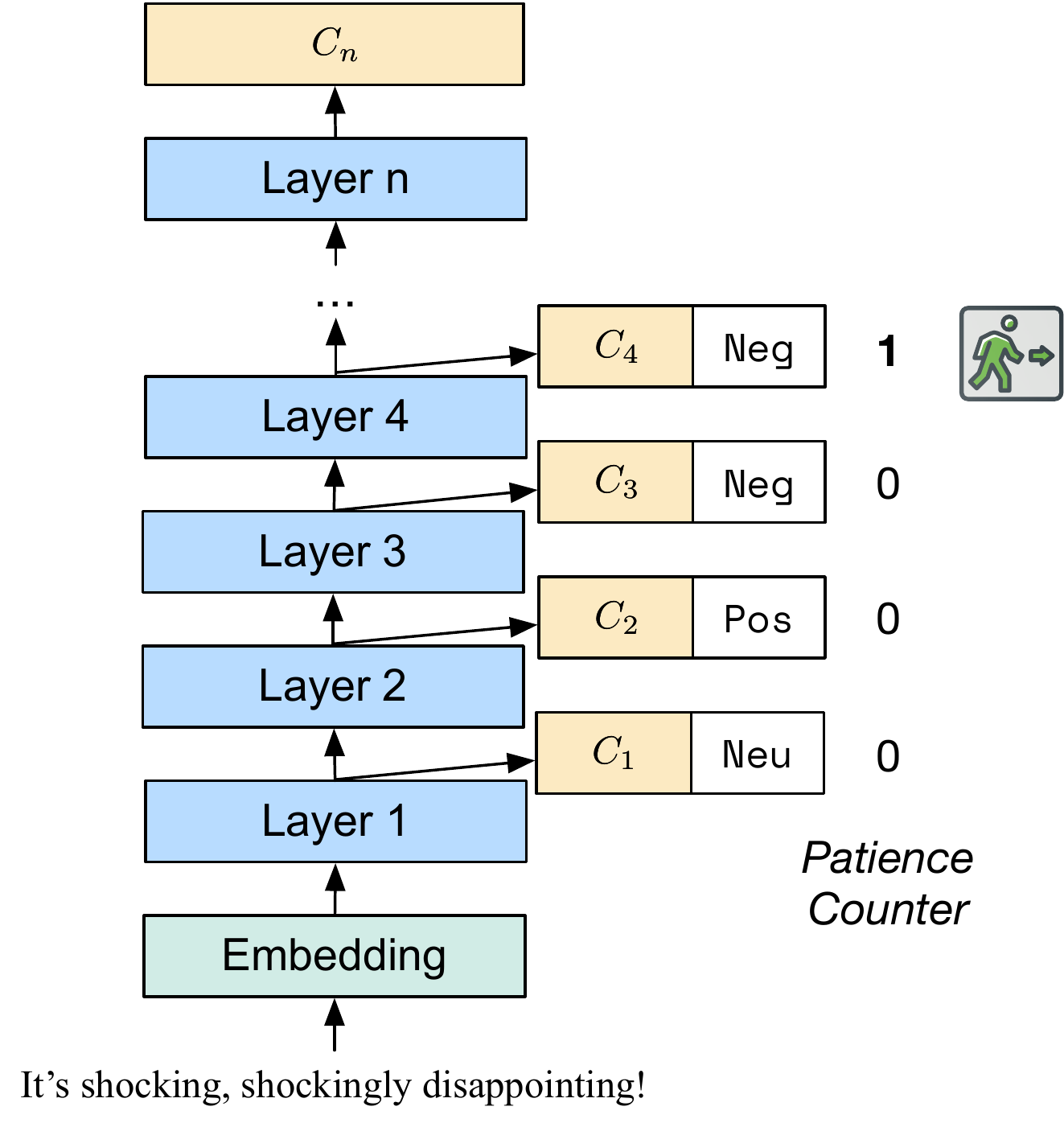}
        \caption{\label{fig:esi}\baby (\abbr)}
    \end{subfigure}
    \caption{\label{fig:cpr}Comparison between Shallow-Deep Net, a prediction score based early exit (threshold is set to $0.9$), and our \baby (patience $t=1$). A classifier is denoted by $C_i$, and $n$ is the number of layers in a model. In this figure, Shallow-Deep incorrectly exits based on the prediction score while \abbr considers multiple classifiers and exits with a correct prediction.}
\end{figure}

In this paper, we propose a novel \textbf{Pa}tience-\textbf{b}ased \textbf{E}arly \textbf{E}xit (\abbr) mechanism to enable models to stop inference dynamically. \abbr is inspired by the widely used Early Stopping~\cite{morgan1990generalization,prechelt1998early} strategy for model training. It enables better input-adaptive inference of PLMs to address the aforementioned limitations.
Specifically, our approach couples an internal classifier with each layer of a PLM and dynamically stops inference when the intermediate predictions of the internal classifiers remain unchanged for $t$ times consecutively (see Figure \ref{fig:esi}), where $t$ is a pre-defined patience. We first show that our method is able to improve the accuracy compared to conventional inference under certain assumptions. Then we conduct extensive experiments on the GLUE benchmark and show that \abbr outperforms existing prediction probability distribution-based exit criteria by a large margin. In addition, \abbr can simultaneously improve inference speed and adversarial robustness of the original model while retaining or even improving its original accuracy with minor additional effort in terms of model size and training time. Also, our method can dynamically adjust the accuracy-efficiency trade-off to fit different devices and resource constraints by tuning the patience hyperparameter without retraining the model, which is favored in real-world applications~\cite{Cai2020Once-for-All}. Although we focus on PLM in this paper, we also have conducted experiments on image classification tasks with the popular ResNet~\cite{he2016deep} as the backbone model and present the results in Appendix \ref{sec:resnet} to verify the generalization ability of \abbr.

To summarize, our contribution is two-fold: (1) We propose \baby, a novel and effective inference mechanism and show its feasibility of improving the efficiency and the accuracy of deep neural networks with theoretical analysis. (2) Our empirical results on the GLUE benchmark highlight that our approach can simultaneously improve the accuracy and robustness of a competitive ALBERT model, while speeding up inference across different tasks with trivial additional training resources in terms of both time and parameters. %

\section{Related Work}

Existing research in improving the efficiency of deep neural networks
can be categorized into two streams: (1) \textit{Static} approaches design compact models or compress heavy models, while the models remain static for all instances at inference (\ie the input goes through the same layers);
(2)
\textit{Dynamic} approaches allow the model to choose different computational paths according to different instances when doing inference. In this way, the simpler inputs usually require less calculation to make predictions. Our proposed \abbr falls into the second category.

\paragraph{Static Approaches: Compact Network Design and Model Compression}

Many lightweight neural network architectures have been specifically designed for resource-constrained applications, including MobileNet~\cite{howard2017mobilenets}, ShuffleNet~\cite{zhang2018shufflenet}, EfficientNet~\cite{tan2019efficientnet}, and ALBERT~\citep{lan2019albert}, to name a few. For model compression, \citet{han2015deep} first proposed to sparsify deep models by removing non-significant synapses and then re-training to restore performance. Weight Quantization~\cite{wu2016quantized} and Knowledge Distillation~\cite{hinton2015distilling} have also proved to be effective for compressing neural models. Recently, existing studies employ Knowledge Distillation~\cite{sanh2019distilbert,sun2019patient,jiao2019tinybert}, Weight Pruning~\cite{michel2019sixteen, voita2019analyzing, fan2019reducing} and Module Replacing~\cite{xu2020bert} to accelerate PLMs.

\paragraph{Dynamic Approaches: Input-Adaptive Inference} 
A parallel line of research for improving the efficiency of neural networks is to enable adaptive inference for various input instances. Adaptive Computation Time~\cite{graves2016adaptive,ut} proposed to use a trainable halting mechanism to perform input-adaptive inference. However, training the halting model requires extra effort and also introduces additional parameters and inference cost. To alleviate this problem, BranchyNet~\cite{teerapittayanon2016branchynet} calculated the entropy of the prediction probability distribution as a proxy for the confidence of branch classifiers to enable early exit. Shallow-Deep Nets~\cite{kaya2018shallow} leveraged the softmax scores of predictions of branch classifiers to mitigate the overthinking problem of DNNs. More recently, \citet{hu2020triple} leveraged this approach in adversarial training to improve the adversarial robustness of DNNs. In addition, existing approaches \cite{graves2016adaptive,wang2018skipnet} trained separate models to determine passing through or skipping each layer. Very recently, FastBERT~\cite{liu2020fastbert} and DeeBERT~\cite{xin2020deebert} adapted confidence-based BranchyNet~\cite{teerapittayanon2016branchynet} for PLMs while RightTool~\cite{Schwartz:2020} leveraged 
the same early-exit criterion as in the Shallow-Deep Network~\cite{kaya2018shallow}.  

However, \citet{Schwartz:2020} recently revealed that prediction probability based methods often lead to substantial performance drop compared to an oracle that identifies the smallest model needed to solve a given instance. In addition, these methods only support classification and leave out regression, which limits their applications. Different from the recent work that directly employs existing efficient inference methods on top of PLMs, \abbr is a novel early-exit criterion that captures the inner-agreement between earlier and later internal classifiers and exploit multiple classifiers for inference, leading to better accuracy.  

\section{\baby}

\begin{figure}
  \centering
  \begin{subfigure}[t]{2.5in}
        \centering
        \includegraphics[width=\textwidth]{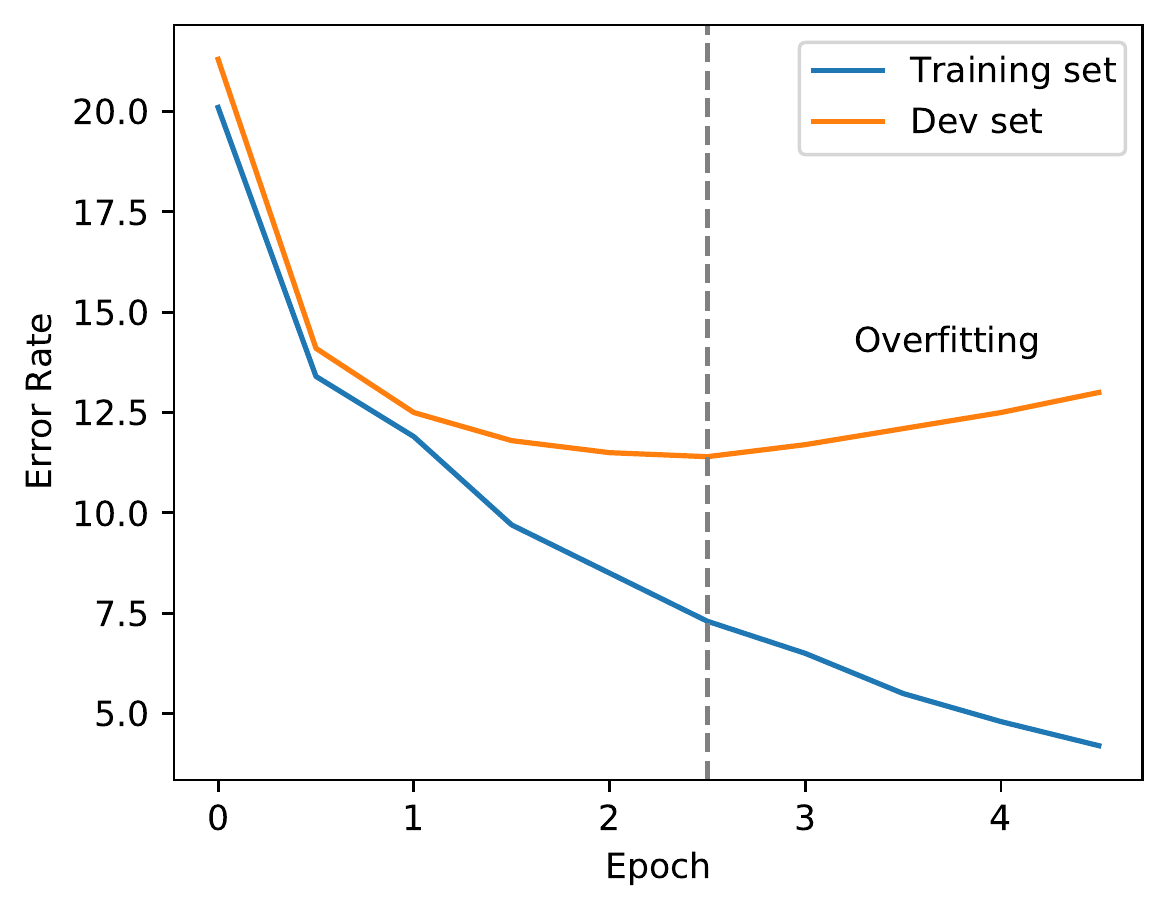}
        \caption{\label{fig:overfit}Overfitting in training}
    \end{subfigure}%
    ~
    \begin{subfigure}[t]{2.8in}
        \centering
        \includegraphics[width=\textwidth]{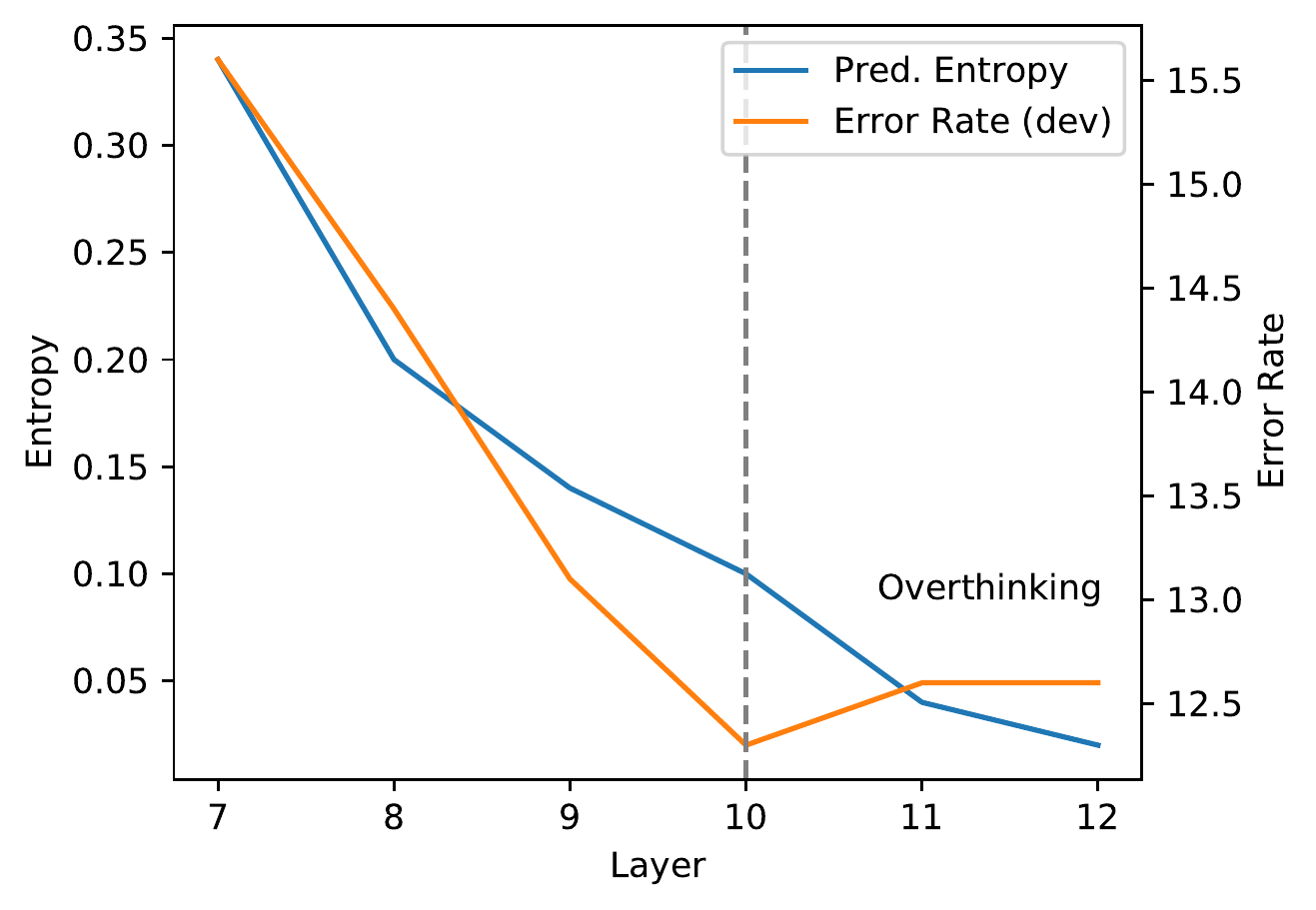}
        \caption{\label{fig:overthink}Overthinking in inference}
    \end{subfigure}
    \caption{\label{fig:analogy}Analogy between overfitting in training and overthinking in inference. \textbf{(a)} In training, the error rate keeps going down on the training set but goes up later on the development set. \textbf{(b)} We insert a classifier after every layer. Similarly, the predicted entropy keeps dropping when more layers are added to inference but the error rate goes up after 10 layers. The results are obtained with ALBERT-base on MRPC.}
\end{figure}

\baby (\abbr) is a plug-and-play method that can work well with minimal adjustment on training.

\subsection{Motivation}

We first conduct experiments to investigate the overthinking problem in PLMs.
As shown in Figure \ref{fig:overthink}, we illustrate the prediction distribution entropy~\cite{teerapittayanon2016branchynet} and the error rate of the model on the development set as more layers join the prediction. Although the model 
becomes
more ``confident'' (lower entropy indicates higher confidence in BranchyNet~\cite{teerapittayanon2016branchynet}) with its prediction as more layers join, the actual error rate instead increases after 10 layers. This phenomenon was discovered and named ``overthinking'' by \citet{kaya2018shallow}. Similarly,
as shown in Figure \ref{fig:overfit}, after 2.5 epochs of training, the model 
continues to get
better accuracy on the training set but begins to deteriorate on the development set. This is the well-known overfitting problem which can be resolved by applying an early stopping mechanism~\cite{morgan1990generalization,prechelt1998early}. From this aspect, overfitting in training and overthinking in inference are naturally alike, inspiring us to adopt an approach similar to early stopping for inference.

\subsection{Inference}
The inference process of \abbr is illustrated in Figure \ref{fig:esi}.
Formally, we define a common inference process as the input instance $\mathbf{x}$ goes through layers $L_1 \ldots L_n$ and the classifier/regressor $C_n$ to predict a class label distribution $\mathbf{y}$ (for classification) or a value $y$ (for regression, we assume the output dimension is $1$ for brevity). We couple an internal classifier/regressor $C_1 \ldots C_{n-1}$ with each layer of $L_1 \ldots L_{n-1}$, respectively. For each layer $L_i$, we first calculate its hidden state $\mathbf{h}_i$:
\begin{equation}
\begin{aligned}
    \mathbf{h}_i & = L_i(\mathbf{h}_{i-1}) \\
    \mathbf{h}_0 & = \operatorname{Embedding}(\mathbf{x})
\end{aligned}
\end{equation}
Then, we use its internal classifier/regressor to output a distribution or value as a per-layer prediction $\mathbf{y}_i = C_i(\mathbf{h}_i)$ or $y_i = C_i(\mathbf{h}_i)$. We use a counter $\mathit{cnt}$ to store the number of times that the predictions remain ``unchanged''. For classification, $\mathit{cnt}_i$ is calculated by:
\begin{equation}
    \mathit{cnt}_i = 
    \begin{cases}
\mathit{cnt}_{i-1} + 1 & \argmax(\mathbf{y}_i) = \argmax(\mathbf{y}_{i-1}),\\
0 & \argmax(\mathbf{y}_i) \neq \argmax(\mathbf{y}_{i-1}) \lor i=0.
\end{cases}
\end{equation}
While for regression, $\mathit{cnt}_i$ is calculated by:
\begin{equation}
    \mathit{cnt}_i = 
    \begin{cases}
\mathit{cnt}_{i-1} + 1 & \left|y_i - y_{i-1}\right| < \tau,\\
0 & \left|y_i - y_{i-1}\right| \geq \tau \lor i=0.
\end{cases}
\end{equation}
where $\tau$ is a pre-defined threshold. We stop inference early at layer $L_j$ when $\mathit{cnt}_j = t$. If this condition is never fulfilled, we use the final classifier $C_n$ for prediction.
In this way, the model can exit early without 
passing
through all layers to make a prediction.

As shown in Figure~\ref{fig:sdn}, prediction score-based early exit relies on the softmax score. As revealed by prior work~\cite{szegedy2013intriguing,jiang2018trust}, prediction of probability distributions (\ie softmax scores) suffers from being over-confident to one class, making it an unreliable metric to represent confidence. Nevertheless, the capability of a low layer may not match its high confidence score. In Figure~\ref{fig:sdn}, the second classifier outputs a high confidence score and incorrectly terminates inference. With \baby, the stopping criteria is in a cross-layer fashion, preventing errors from one single classifier. Also, since \abbr comprehensively considers results from multiple classifiers, it can also benefit from an ensemble learning~\cite{krogh1994ensemble} effect.%

\subsection{Training}
\label{training}
\abbr requires that we
train internal classifiers to predict based on their corresponding layers' hidden states. 
For classification, the loss function $\mathcal{L}_i$ for classifier $C_i$ is calculated with cross entropy:
\begin{equation}
    \mathcal{L}_i=-\sum_{z \in Z}\left[\mathbbm{1}\left[\mathbf{y}_i=z\right]\cdot\log P\left(\mathbf{y}_i=z|\mathbf{h}_i\right)\right]
\end{equation}
where $z$ and $Z$ denote a class label and the set of class labels, respectively.
For regression, the loss is instead calculated by a (mean) squared error:
\begin{equation}
    \mathcal{L}_i = \left(y_i-\hat{y}_i\right)^2
\end{equation}
where $\hat{y}$ is the ground truth. Then, we calculate and train the model to minimize the total loss $\mathcal{L}$ by a weighted average following~\citet{kaya2018shallow}:
\begin{equation}
    \mathcal{L} = \frac{\sum_{j=1}^{n} j\cdot\mathcal{L}_j}{\sum_{j=1}^{n}j}
\end{equation}
In this way, every possible inference branch has been covered in the training process. Also, the weighted average can correspond to the relative inference cost of each internal classifier.

\subsection{Theoretical Analysis}

It is straightforward to see that 
\baby 
is able to reduce inference latency. To understand whether and under what conditions it can also improve accuracy, we conduct a theoretical comparison of a model's accuracy with and without \abbr under a simplified condition.
We consider the case of binary classification for simplicity and conclude that:
\begin{theorem}
\label{th1}
Assuming the patience of \abbr inference is $t$, the total number of internal classifiers (IC) is $n$, the misclassification probability (\ie error rate) of all internal classifiers (excluding the final classifier) is $q$, and the misclassification probability of the final classifier and the original classifier (without ICs) is $p$. Then the \abbr mechanism improves the accuracy of conventional inference as long as $n-t < (\frac{1}{2q})^{t}(\frac{p}{q}) -p $
\end{theorem}
(the proof is detailed in Appendix \ref{sec:proof}).

We can see the above inequality can be easily satisfied. For instance, when $n=12$, $q=0.2$, and $p=0.1$, the above equation is satisfied as long as the patience $t\geq4$. However, it is notable that assuming the accuracy of each internal classifiers to be equal and independent is generally not attainable in practice. Additionally, we verify the statistical feasibility of \abbr with Monte Carlo simulation in Appendix \ref{sec:montecarlo}. To further test \abbr with real data and tasks, we also conduct extensive experiments in the following section.

\section{Experiments}

\subsection{Tasks and Datasets}

We evaluate our proposed approach on the GLUE benchmark~\cite{glue}. Specifically, we test on Microsoft Research Paraphrase Matching (MRPC)~\cite{mrpc}, Quora Question Pairs (QQP)\footnote{\url{https://www.quora.com/q/quoradata/First-Quora-Dataset-Release-Question-Pairs}} and STS-B~\cite{senteval} for Paraphrase Similarity Matching; Stanford Sentiment Treebank (SST-2)~\cite{sst} for Sentiment Classification; Multi-Genre Natural Language Inference Matched (MNLI-m), Multi-Genre Natural Language Inference Mismatched (MNLI-mm)~\cite{mnli}, Question Natural Language Inference (QNLI)~\cite{qnli} and Recognizing Textual Entailment (RTE)~\cite{glue} for the Natural Language Inference (NLI) task; The Corpus of Linguistic Acceptability (CoLA)~\cite{cola} for Linguistic Acceptability. We exclude WNLI~\cite{wnli} from GLUE following previous work~\cite{devlin2018bert,jiao2019tinybert,xu2020bert}. For datasets with more than one metric, we report the arithmetic mean of the metrics.

\subsection{Baselines}

For GLUE tasks, we compare our approach with four types of baselines: \textbf{(1) Backbone models:} We choose ALBERT-base and BERT-base, which have approximately the same inference latency and accuracy. \textbf{(2) Directly reducing layers:} We experiment with the first 6 and 9 layers of the original (AL)BERT with a single output layer on the top, denoted by (AL)BERT-6L and (AL)BERT-9L, respectively. These two baselines help to set a lower bound 
for methods
that do not employ any technique.
\textbf{(3) Static model compression approaches:} For pruning, we include the results of LayerDrop~\cite{fan2019reducing} and attention head pruning~\cite{michel2019sixteen} on ALBERT. For reference, we also report the performance of state-of-the-art methods on compressing the BERT-base model with knowledge distillation or module replacing, including DistillBERT~\cite{sanh2019distilbert}, BERT-PKD~\cite{sun2019patient} and BERT-of-Theseus~\cite{xu2020bert}. 
\textbf{(4) Input-adaptive inference:} Following the settings in concurrent studies~\cite{Schwartz:2020,liu2020fastbert,xin2020deebert}, we add internal classifiers after each layer and apply different early exit criteria, including that employed by BranchyNet~\cite{teerapittayanon2016branchynet} and Shallow-Deep~\cite{kaya2018shallow}. To make a fair comparison, the internal classifiers and their insertions are exactly same in both baselines and \baby. We search over a set of thresholds to find the one delivering the best accuracy for the baselines while targeting a speed-up ratio between $1.30\times$ and $1.96\times$ (the speed-up ratios of (AL)BERT-9L and -6L, respectively).

\subsection{Experimental Setting}

\paragraph{Training} We add a linear output layer after each intermediate layer of the pretrained BERT/ALBERT model as the internal classifiers. We perform grid search over batch sizes of \{16, 32, 128\}, and learning rates of \{1e-5, 2e-5, 3e-5, 5e-5\} with an Adam optimizer. We apply an early stopping mechanism and select the model with the best performance on the development set. We implement \abbr on the base of Hugging Face's Transformers~\cite{wolf2020huggingfaces}. We conduct our experiments on a single Nvidia V100 16GB GPU. 

\paragraph{Inference} Following prior work on input-adaptive inference~\cite{teerapittayanon2016branchynet,kaya2018shallow}, inference is on a per-instance basis, \ie the batch size for inference is set to 1. This is a common latency-sensitive production scenario when processing individual requests from different users~\cite{Schwartz:2020}. We report the median performance over 5 runs with different random seeds because the performance on relatively small datasets such as CoLA and RTE usually has large variance. For \abbr, we set the patience $t=6$ in the overall comparison to keep the speed-up ratio between $1.30\times$ and $1.96\times$ while obtaining good performance 
following
Figure \ref{fig:patience}. We further analyze the behavior of the \abbr mechanism with different patience settings in Section \ref{sec:patience}.

\begin{table*}[tb]
\caption{Experimental results (median of 5 runs) of models with ALBERT backbone on the development set and the test set of GLUE. The numbers under each dataset indicate the number of training samples. The acceleration ratio is averaged across 8 tasks. We mark ``-'' on STS-B for BranchyNet and Shallow-Deep since they do not support regression.}
\label{tab:main}
\vskip 0.1in
\begin{center}
\resizebox{1.\linewidth}{!}{
\begin{tabular}{l|cc|cccccccc|c}
\toprule
\multirow{2}{*}{\textbf{Method}} & \multirow{2}{*}{\textbf{\#Param}} & \textbf{Speed} & \textbf{CoLA} & \textbf{MNLI} & \textbf{MRPC} & \textbf{QNLI} & \textbf{QQP} & \textbf{RTE} & \textbf{SST-2} & \textbf{STS-B} & \textbf{Macro} \\
& & \textbf{-up} & (8.5K) & (393K) & (3.7K) & (105K) & (364K) & (2.5K) & (67K) & (5.7K) & \bf Score\\
\midrule
\multicolumn{12}{c}{\textit{Dev. Set}} \\
\midrule
ALBERT-base~\cite{lan2019albert}  & 12M & 1.00$\times$ & 58.9 & 84.6 & 89.5 & 91.7 & 89.6 & 78.6 & 92.8 & 89.5 & 84.4 \\
\midrule
ALBERT-6L & 12M & 1.96$\times$  & 53.4 & 80.2 & 85.8 & 87.2 & 86.8 & 73.6 & 89.8 & 83.4 & 80.0 \\
ALBERT-9L & 12M & 1.30$\times$ & 55.2 & 81.2 & 87.1 & 88.7 & 88.3 & 75.9 & 91.3 & 87.1 & 81.9 \\
\midrule
LayerDrop~\cite{fan2019reducing} & 12M & 1.96$\times$ & 53.6 & 79.8 & 85.9 & 87.0 & 87.3 & 74.3 & 90.7 & 86.5 & 80.6 \\
HeadPrune~\cite{michel2019sixteen} & 12M & 1.22$\times$ & 54.1 & 80.3 & 86.2 & 86.8 & 88.0 & 75.1 & 90.5 & 87.4 & 81.1 \\
\midrule
BranchyNet~\cite{teerapittayanon2016branchynet} & 12M & 1.88$\times$ & 55.2 & 81.7 & 87.2 & 88.9 & 87.4 & 75.4 & 91.6 & - & - \\
Shallow-Deep~\cite{kaya2018shallow} & 12M & 1.95$\times$ & 55.5 & 81.5 & 87.1 & 89.2 & 87.8 & 75.2 & 91.7 & - & - \\
\abbr \textit{(ours)} & 12M & 1.57$\times$ & \textbf{61.2} & \textbf{85.1} & \textbf{90.0} & \textbf{91.8} & \textbf{89.6} & \textbf{80.1} & \textbf{93.0} & \textbf{90.1} & \textbf{85.1} \\
\midrule
\multicolumn{12}{c}{\textit{Test Set}} \\
\midrule
ALBERT-base~\cite{lan2019albert}  & 12M & 1.00$\times$ & 54.1 & 84.3 & 87.0 & 90.8 & 71.1 & 76.4 & 94.1 & 85.5 & 80.4 \\
\abbr \textit{(ours)} & 12M & 1.57$\times$ & \textbf{55.7} & \textbf{84.8} & \textbf{87.4} & \textbf{91.0} & \textbf{71.2} & \textbf{77.3} & \textbf{94.1} & \textbf{85.7} &\textbf{80.9} \\
\bottomrule
\end{tabular}}
\end{center}
\vskip -0.1in
\end{table*}

 \begin{table}[!ht]
 \RawFloats
    \centering
\begin{minipage}[t]{0.58\linewidth}\centering
\caption{\label{tab:bert}Experimental results (median of 5 runs) of BERT based models on the development set of GLUE. We mark ``-'' on STS-B for BranchyNet and Shallow-Deep since they do not support regression.}
\resizebox{1.\linewidth}{!}{
 \begin{tabular}{l|cc|ccc}
\toprule
\multirow{2}{*}{\textbf{Method}} & \multirow{2}{*}{\textbf{\#Param}} & \textbf{Speed} & \textbf{MNLI} & \textbf{SST-2} & \textbf{STS-B} \\
& & \textbf{-up} & (393K) & (67K) & (5.7K) \\
\midrule
BERT-base~\cite{devlin2018bert}  & 108M & 1.00$\times$  & 84.5 & 92.1 & 88.9 \\
\midrule
BERT-6L & 66M & 1.96$\times$  & 80.1 & 89.6 & 81.2 \\
BERT-9L & 87M & 1.30$\times$ & 81.4 & 90.5 & 85.0 \\
\midrule
DistilBERT~\cite{sanh2019distilbert} & 66M & 1.96$\times$ & 79.0 & 90.7 & 81.2 \\
BERT-PKD~\cite{xu2020bert} & 66M & 1.96$\times$ & 81.3 & 91.3 & 86.2 \\
BERT-of-Theseus~\cite{xu2020bert} & 66M & 1.96$\times$ & 82.3 & 91.5 & \textbf{88.7} \\
\midrule
BranchyNet~\cite{teerapittayanon2016branchynet} & 108M & 1.87$\times$ & 80.3 & 90.4 & - \\
Shallow-Deep~\cite{kaya2018shallow} & 108M & 1.91$\times$ & 80.5 & 90.6 & - \\
\abbr \textit{(ours)} & 108M & 1.62$\times$ & \textbf{83.6} & \textbf{92.0} & \bf 88.7\\
\bottomrule
\end{tabular}
}
\end{minipage}\hfill%
\begin{minipage}[t]{0.4\linewidth}\centering
\caption{Parameter numbers and training time (in minutes) until the best performing checkpoint (on the development set) with and without \abbr on ALBERT and BERT as backbone models.}
\label{tab:trainingtime}
\resizebox{\linewidth}{!}{
\begin{tabular}{l|cc|cc}
\toprule
\multirow{2}{*}{Method}    & \multicolumn{2}{c|}{\#Param} & \multicolumn{2}{c}{Train. time (min)} \\
& MNLI & SST-2 & MNLI & SST-2\\
\midrule
\multicolumn{4}{l}{\textbf{ALBERT}}\\
\midrule
w/o \abbr & 12M & 12M & 234  & 113 \\
w/ \abbr  & +36K & +24K & \textbf{227}  & \textbf{108}  \\
\midrule
\multicolumn{4}{l}{\textbf{BERT}}\\
\midrule
w/o \abbr & 108M & 108M & 247 & 121 \\
w/ \abbr  & +36K & +24K & \textbf{242}  & \textbf{120} \\
\bottomrule
\end{tabular}
}
\end{minipage}
\end{table}

\begin{sbsfigure}
\RawFloats
    \begin{minipage}[t]{\textwidth}
  \begin{minipage}[t]{0.49\textwidth}
    \centering
\includegraphics[width=\textwidth]{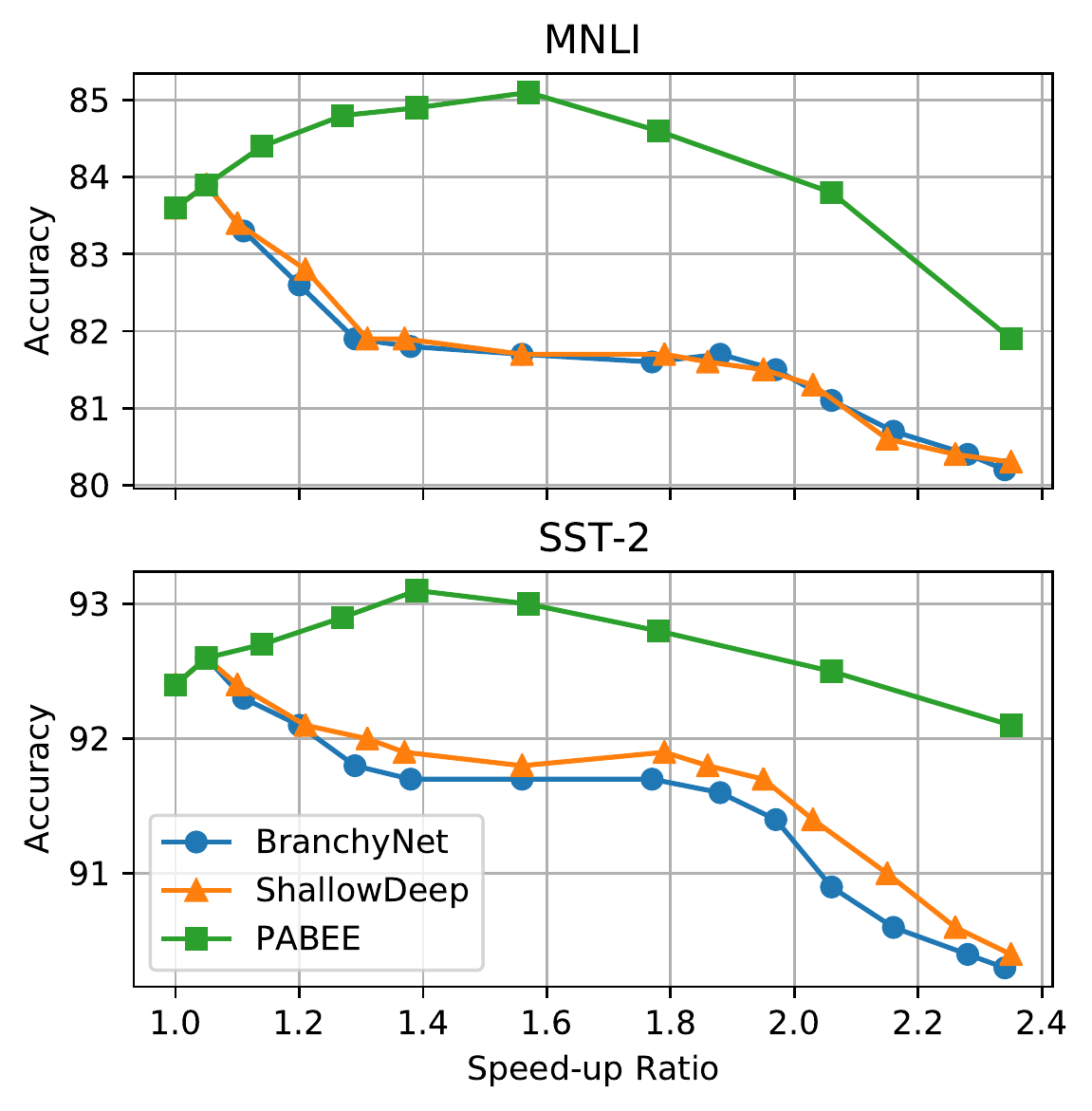}
    \captionof{figure}{\label{fig:tradeoff}Speed-accuracy curves of BranchyNet, Shallow-Deep and \abbr on MNLI and SST-2 with ALBERT-base model.}
    \end{minipage}
    \hfill
\begin{minipage}[t]{0.49\textwidth}
  \centering
\includegraphics[width=\textwidth]{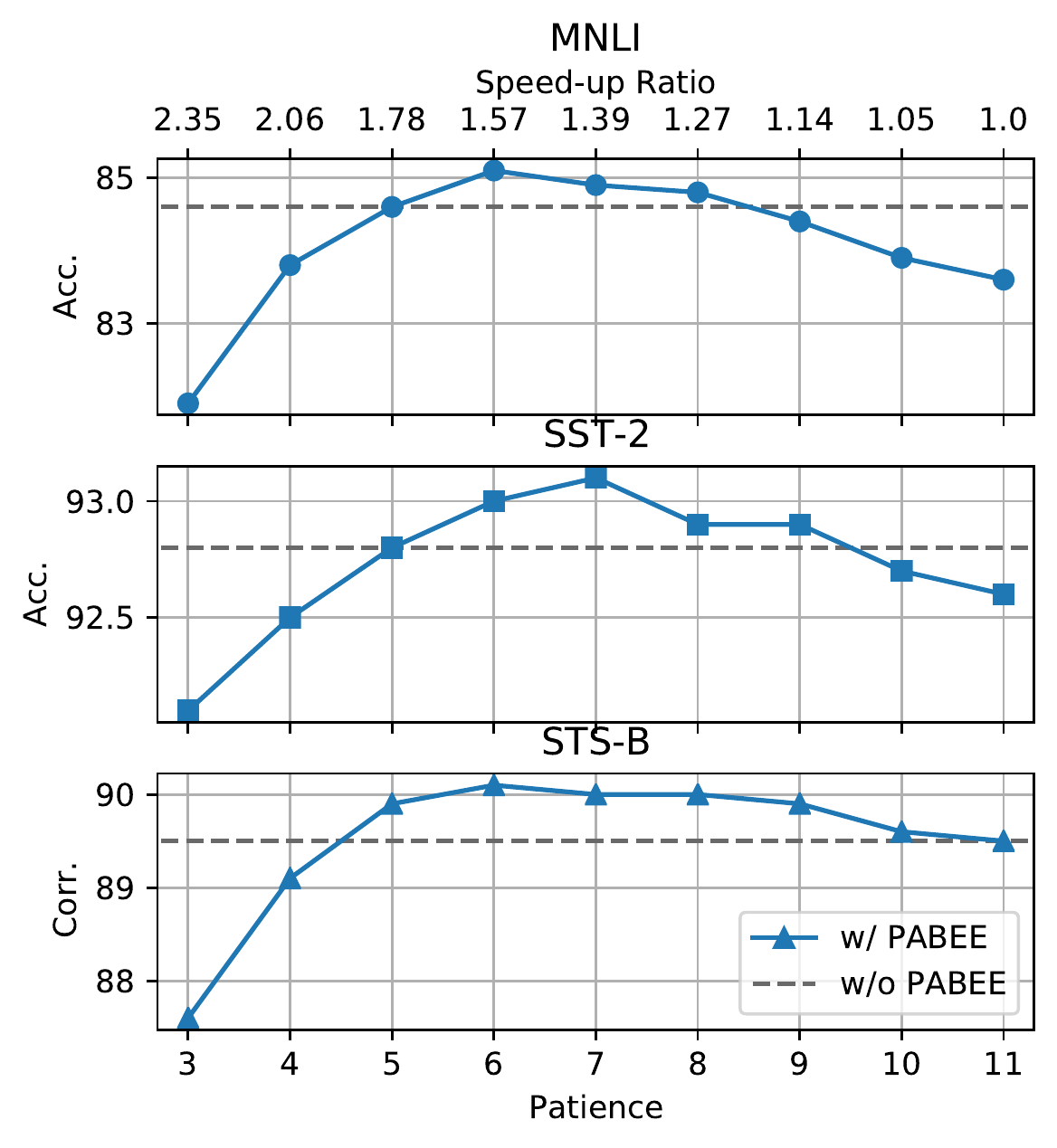}
    \captionof{figure}{\label{fig:patience}Accuracy scores and speed-up ratios under different patience with ALBERT-base model. The baseline is denoted with gray dash lines.}
  \end{minipage}
  \end{minipage}
\end{sbsfigure}

\subsection{Overall Comparison}

We first report our main result on GLUE with ALBERT as the backbone model in Table \ref{tab:main}. This choice is made because: (1) ALBERT is a state-of-the-art PLM for natural language understanding. (2) ALBERT is already very efficient in terms of the number of parameters and memory use because of its layer sharing mechanism, but still suffers from the problem of high inference latency. We can see that our approach outperforms all compared approaches on improving the inference efficiency of PLMs, demonstrating the effectiveness of the proposed \abbr mechanism. Surprisingly, our approach consistently improves the performance of the original ALBERT model by a relatively large margin while speeding-up inference 
by $1.57 \times$. 
This is, to the best of our knowledge, the first inference strategy that can improve both the speed and performance of a fine-tuned PLM. 

To better compare the efficiency of \abbr with the method employed in BranchyNet and Shallow-Deep, we illustrate speed-accuracy curves in Figure \ref{fig:tradeoff} with different trade-off hyperparameters (\ie threshold for BranchyNet and Shallow-Deep, patience for \abbr). Notably, \abbr 
retains higher accuracy than BranchyNet and Shallow-Deep under the same speed-up ratio, showing its superiority over prediction score based methods. 

To demonstrate the versatility of our method with different PLMs, we report the results on a representative subset of GLUE with BERT~\cite{devlin2018bert} as the backbone model in Table \ref{tab:bert}. We can see that our BERT-based model significantly outperforms other compared methods with either knowledge distillation or prediction probability based input-adaptive inference methods. Notably, the performance is slightly lower than the original BERT model while \abbr improves the accuracy on ALBERT. We suspect that this is because the intermediate layers of BERT have never been connected to an output layer during pretraining, which leads to a mismatch between pretraining and fine-tuning when adding the internal classifiers. 
However, \abbr still has a higher accuracy than various knowledge distillation-based approaches as well as prediction probability distribution based models, showing its potential as a generic method for deep neural networks of different kinds.

As for the cost of training, we present parameter numbers and training time with and without \abbr with both BERT and ALBERT backbones in Table \ref{tab:trainingtime}. Although more classifiers need to be trained, training \abbr is no slower (even slightly faster) than conventional fine-tuning, which may be attributed to the additional loss functions of added internal classifiers. This makes our approach appealing compared with other approaches for accelerating inference such as pruning or distillation because they require separately training another model for each speed-up ratio in addition to training the full model. Also, \abbr only introduces fewer than 40K parameters ($0.33\%$ of the original 12M parameters).

\subsection{Analysis}

\paragraph{Impact of Patience}
\label{sec:patience}
As illustrated in Figure \ref{fig:patience}, different patience can lead to different speed-up ratios and performance. For a 12-layer ALBERT model, \abbr reaches peak performance with a patience of 6 or 7. On MNLI, SST-2 and STS-B, \abbr can always outperform the baseline with patience between 5 and 8. Notably, unlike BranchyNet and Shallow-Deep, whose accuracy drops as the inference speed goes up, \abbr has an inverted-U curve. We confirm this observation statistically with Monte Carlo simulation in Appendix \ref{sec:montecarlo}. To analyze, when the patience $t$ is set too large, the later internal classifier may suffer from the overthinking problem and make a wrong prediction that breaks the stable state among previous internal classifiers, which have not met the early-exit criterion because $t$ is large. This makes \abbr leave more samples to be classified by the final classifier $C_n$, which suffers from the aforementioned overthinking problem. Thus, the effect of the ensemble learning vanishes and undermines its performance. Similarly, when $t$ is relatively small, more samples may meet the early-exit criterion by accident before actually reaching the stable state where consecutive internal classifiers agree with each other.

\paragraph{Impact of Model Depth}
\label{sec:depth}

We also investigate the impact of model depth on the performance of \abbr. We apply \abbr to a 24-layer ALBERT-large model. As shown in Table \ref{tab:depth}, our approach consistently improves the accuracy as more layers and classifiers are added while producing an even larger speed-up ratio. This finding demonstrates the potential of \abbr for burgeoning deeper PLMs~\cite{shoeybi2019megatron,raffel2019exploring,brown2020language}.

\begin{table*}[tb]
\resizebox{0.7\linewidth}{!}{
 \begin{tabular}{l|ccc|ccc}
\toprule
\multirow{2}{*}{\textbf{Method}} & \multirow{2}{*}{\textbf{\#Param}} & \multirow{2}{*}{\textbf{\#Layer}} & \textbf{Speed} & \textbf{MNLI} & \textbf{SST-2} & \textbf{STS-B} \\
& & & \textbf{-up} & (393K) & (67K) & (5.7K) \\
\midrule
ALBERT-base~\cite{lan2019albert}  & 12M & 12 & 1.00$\times$ & 84.6 & 92.8 & 89.5 \\
+ \abbr & 12M & 12 & 1.57$\times$ & \textbf{85.1} & \textbf{93.0} & \textbf{90.1} \\
\midrule
ALBERT-large~\cite{lan2019albert}  & 18M & 24 & 1.00$\times$ & 86.4 & 94.9 & 90.4 \\
+ \abbr & 18M & 24 & 2.42$\times$ & \textbf{86.8} & \textbf{95.2} & \textbf{90.6}\\

\bottomrule
\end{tabular}
}
 \caption{\label{tab:depth}Experimental results (median of 5 runs) of different sizes of ALBERT on GLUE development set.}

\end{table*}

\begin{table*}[t]
\caption{Results on the adversarial robustness. ``Query Number'' denotes the number of queries the attack system made to the target model and a higher number indicates more difficulty. }
\label{tab:robustness}
\begin{center}
\resizebox{\textwidth}{!}{
\begin{tabular}{l|ccc|ccc|ccc}
\toprule
\textbf{Metric} & \multicolumn{3}{|c|}{\textbf{ALBERT}} & \multicolumn{3}{|c}{\textbf{+ Shallow-Deep~\cite{kaya2018shallow}}} & \multicolumn{3}{|c}{\textbf{+ \abbr} \textit{(ours)}} \\
 ($\uparrow$ better) & SNLI & MNLI-m/-mm & Yelp & SNLI & MNLI-m/-mm & Yelp & SNLI & MNLI-m/-mm & Yelp \\
\midrule
\textbf{Original Acc.} & 89.6 & 84.1 / 83.2 & 97.2 & 89.4 & 82.2 / 80.5 & 97.2 & \bf 89.9 & \bf 85.0 / 84.8 & \bf 97.4 \\
\textbf{After-Attack Acc.} & 5.5 & 9.8 / 7.9 & 7.3 & 9.2 & 15.4 / 13.8 & 11.4 & \bf 19.3 & \bf 30.2 / 25.6 & \bf 18.1 \\
\textbf{Query Number} & 58 & 80 / 86 & 841 & 64 & 82 / 86 & 870 & \bf 75 & \bf 88 / 93 & \bf 897 \\
\bottomrule
\end{tabular}}
\end{center}
\vskip -0.3in
\end{table*}

\subsection{Defending
Against
Adversarial Attack}
\label{robust}

Deep Learning models have been found to be vulnerable to adversarial examples that are slightly altered with perturbations often indistinguishable to humans~\cite{kurakin2017adversarial}. \citet{jin2019bert} revealed that PLMs can also be attacked with a high success rate. Recent studies~\cite{kaya2018shallow,hu2020triple} attribute the vulnerability partially to the overthinking problem, arguing that it can be mitigated by early exit mechanisms.

In our experiments, we use a state-of-the-art adversarial attack method, TextFooler~\cite{jin2019bert}, which demonstrates effectiveness on attacking BERT. We conduct black-box attacks on three datasets: SNLI~\cite{snli}, MNLI~\cite{mnli} and Yelp~\cite{yelp}. Note that since we use the pre-tokenized data provided by \citet{jin2019bert}, the results on MNLI differ slightly from the ones in Table~\ref{tab:main}. We attack the original ALBERT-base model, ALBERT-base with Shallow-Deep~\cite{kaya2018shallow} and with \baby.

As shown in Table~\ref{tab:robustness}, we report the original accuracy, after-attack accuracy and the number of queries needed by TextFooler to attack each model. Our approach successfully defends more than $3\times$ attacks compared to the original ALBERT on NLI tasks, and $2\times$ on the Yelp sentiment analysis task. Also, \abbr increases the number of queries needed to attack by a large margin, providing more protection to the model. Compared to Shallow-Deep~\cite{kaya2018shallow}, our model demonstrates significant robustness improvements. To analyze, although the early exit mechanism of Shallow-Deep can prevent the aforementioned overthinking problem, it still 
relies on a single classifier to make the final prediction, which makes it %
vulnerable to adversarial attacks. %
In comparison,
since \baby exploits multiple layers and classifiers, the attacker has to fool multiple classifiers (which may exploit different features) at the same time, making it much more difficult to attack the model. This effect is similar to the merits of ensemble learning against adversarial attack, discussed in previous studies~\cite{strauss2017ensemble,tramer2018ensemble,pang2019improving}. 

\section{Discussion}
In this paper, we proposed \abbr, a novel efficient inference method that can yield better accuracy-speed trade-off than existing methods. We verify its effectiveness and efficiency on GLUE and provide theoretical analysis. Empirical results show that \abbr can simultaneously improve the efficiency, accuracy, and adversarial robustness upon a competitive ALBERT model. However, a limitation is that \abbr currently only works on models with a single branch (\eg ResNet, Transformer). Some adaption is needed for multi-branch networks (\eg NASNet~\cite{zoph2018learning}).
For future work, we would like to explore our method on more tasks and settings. Also, since \abbr is 
orthogonal to prediction distribution based early exit approaches, it would be interesting to see if we can combine them with \abbr for better performance.

\section*{Broader Impact}
As an efficient inference technique, our proposed \abbr can facilitate more applications on mobile and edge computing, and also help reduce energy use and carbon emission~\cite{schwartz2019green}.
Since our method serves as a plug-in for existing pretrained language models, it does not introduce significant new ethical concerns but more work is needed to determine its effect on biases (\eg gender bias) that have already been encoded in a PLM.

\begin{ack}
We are grateful for the comments from the anonymous reviewers. We would like to thank the authors of TextFooler~\cite{jin2019bert}, Di Jin and Zhijing Jin, for their help with the data for adversarial attack. Tao Ge is the corresponding author.
\end{ack}

\bibliography{ref}{}
\bibliographystyle{unsrtnat}

\newpage
\appendix

\section{Image Classification}
\label{sec:resnet}

To verify the effectiveness of \abbr on Computer Vision, we follow the experimental settings in Shallow-Deep~\cite{kaya2018shallow},
we conduct experiments on two image classification datasets, CIFAR-10 and CIFAR-100~\cite{krizhevsky2009learning}. We use ResNet-56~\cite{he2016deep} as the backbone and compare \abbr with BranchyNet~\cite{teerapittayanon2016branchynet} and Shallow-Deep~\cite{kaya2018shallow}. After every two convolutional layers, an internal classifier is added. We set the batch size to 128 and use SGD optimizer with learning rate of $0.1$.
 
 \begin{table}[h]
     \centering
 \begin{tabular}{l|cc|cc}
\toprule
\multirow{2}{*}{\textbf{Method}} & \multicolumn{2}{c}{\textbf{CIFAR-10}} & \multicolumn{2}{c}{\textbf{CIFAR-100}} \\
& Speed-up & Acc. & Speed-up & Acc. \\
\midrule
ResNet-56~\cite{he2016deep}  & 1.00$\times$  & 91.8 & 1.00$\times$  & 68.6 \\
\midrule
BranchyNet~\cite{teerapittayanon2016branchynet}  & 1.33$\times$  & 91.4 & 1.29$\times$  & 68.2 \\
Shallow-Deep~\cite{kaya2018shallow}  & 1.35$\times$  & 91.6 & 1.32$\times$  & 68.8 \\
\abbr \textit{(ours)} & 1.26$\times$  & \bf 92.0 & 1.22$\times$  & \bf 69.1 \\
\bottomrule
\end{tabular}
 \caption{\label{tab:cifar}Experimental results (median of 5 runs) of ResNet based models on CIFAR-10 and CIFAR-100 datasets.}
 \end{table}

The experimental results in CIFAR are reported in Table \ref{tab:cifar}. \abbr outperform the original ResNet model by $0.2$ and $0.5$ in terms of accuracy while speed up the inference by $1.26\times$ and $1.22\times$ on CIFAR-10 and CIFAR-100, respectively. Also, \abbr demonstrates a better performance and a similar speed-up ratio compared to both baselines.

\section{Proof of Theorem \ref{th1}}
\label{sec:proof}

\begin{proof}

Recap we are in the case of binary classification. We denote the patience of \abbr as $t$, the total number of internal classifiers (IC) as $n$, the misclassification probability (\ie error rate) of all internal classifiers as $q$, and the misclassification probability of the final classifier and the original classifier as $p$. We want to prove the \abbr mechanism improves the accuracy of conventional inference as long as $n-t < (\frac{1}{2q})^{t+1}p -q $.

For the examples that do not early-stopped, the misclassification probability with and without \abbr is the same. Therefore, we only need to consider the ratio between the probability that a sample is early-stopped and misclassified (denoted as $p_\mathit{misc}$) and that a sample is early-stopped (denoted as $p_\mathit{stop}$). We want to find the condition on $n$ and $t$ which makes $\frac{p_\mathit{misc}}{p_\mathit{stop}} < p $.

First, considering only the probability mass of the model consecutively output the same label from the first position, we have
\begin{equation}
    p_\mathit{stop} > q^{t+1} + (1-q)^{t+1}
\end{equation}
which is the lower bound of $p_\mathit{stop}$ that only considering the probability of a sample is early-stopped by consecutively predicted to be the same label from the first internal classifier. We then take its derivative and find it obtains its minimum when $q=0.5$. This corresponds to the case where the classification is performing random guessing (i.e. classification probability for class 0 and 1 equal to 0.5). Intuitively, in the random guessing case the internal classification results are most instable so the probability that a sample is early-stopped is the smallest.
 
Therefore, we have $p_\mathit{stop} > (\frac{1}{2})^{t}$.

Then for $p_\mathit{misc}$, we have 

\begin{equation}
    p_{misc} < q^{t+1} + (n-t-1)(1-q)q^{t+1}
\end{equation}

where $q^{t+1}$ is the probability that the example is consecutively misclassified for t+1 times from the first IC. The term $(n-t-1)(1-q)q^{t+1}$ is the summation of probability that the example is consecutively misclassified for t+1 times from the $2, ..., n-t$ th IC but correctly classified at the previous IC, without considering the cases that the the inference may already finished (whether correctly or not) before that IC. The summation of these two terms is an upper bound of $p_{misc}$.

So we need to have 

\begin{equation}
    (n-t)q^{t+1} - (n-m-1)q^{t+2} < (\frac{1}{2})^{t} p
\end{equation}

which equals to 

\begin{equation}
    (n-t)(q^{t}-q^{t+1}) < (\frac{1}{2})^{t}(\frac{p}{q}) - q^{t+1}
\end{equation}

which equals to 

\begin{equation}
    n-t < \frac{(\frac{1}{2q})^{t}(\frac{p}{q}) - q}{1-q} < (\frac{1}{2q})^{t}(\frac{p}{q}) - q
\end{equation}
$\hfill\blacksquare$ 
\end{proof}

Specially, when $q = p$, the condition becomes $n-t < (\frac{1}{2p})^{t} -p $

\section{Monte Carlo Simulation}
\label{sec:montecarlo}
To verify the theoretical feasibility of \baby, we conduct Monte Carlo simulation. We simplify the task to be a binary classification with a 12-layer model which has classifiers $C_1 \ldots C_{12}$ that all have the same probability to correctly predict.

\begin{figure}[h]
  \centering
  \begin{subfigure}[t]{0.45\textwidth}
        \centering
        \includegraphics[width=\textwidth]{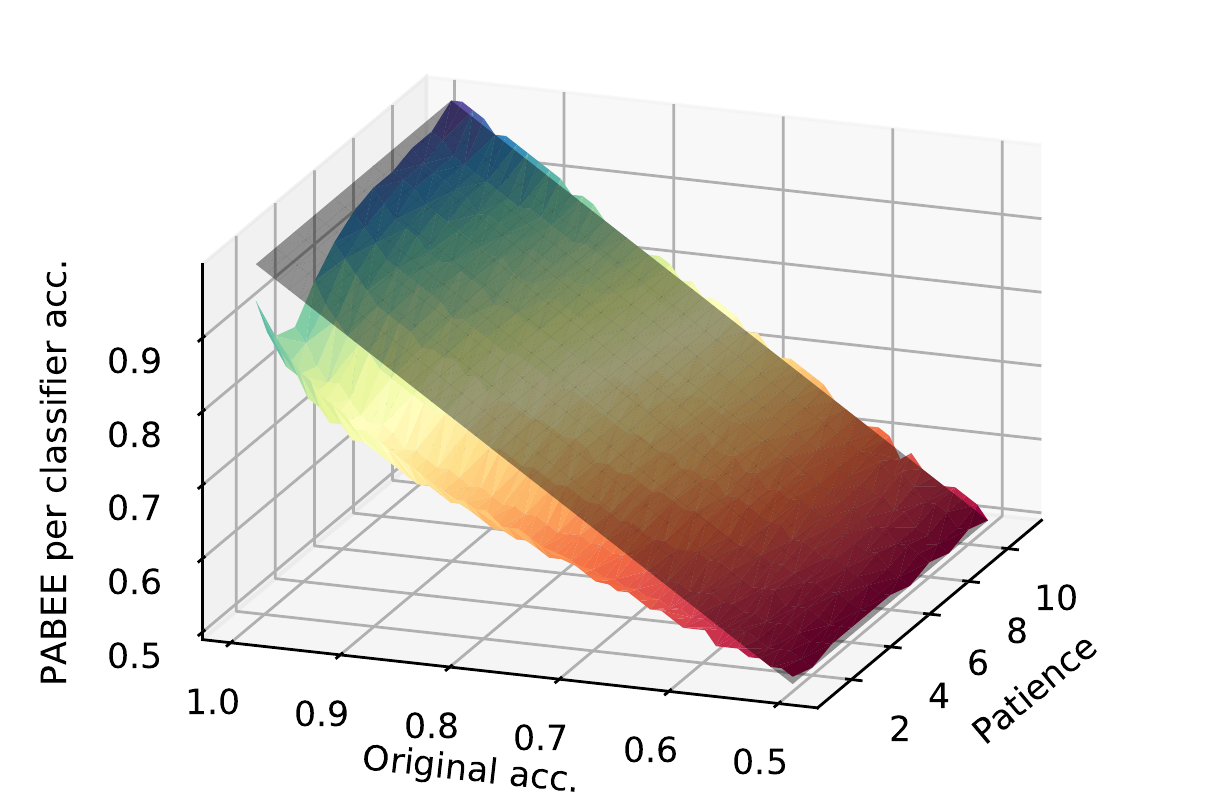}
        \caption{\label{fig:acclb}Accuracy lower bound of each single \abbr classifier to achieve the original accuracy. The translucent black plain denotes inference without \abbr.}
    \end{subfigure}%
    ~\hspace{0.2in}
    \begin{subfigure}[t]{0.45\textwidth}
        \centering
        \includegraphics[width=\textwidth]{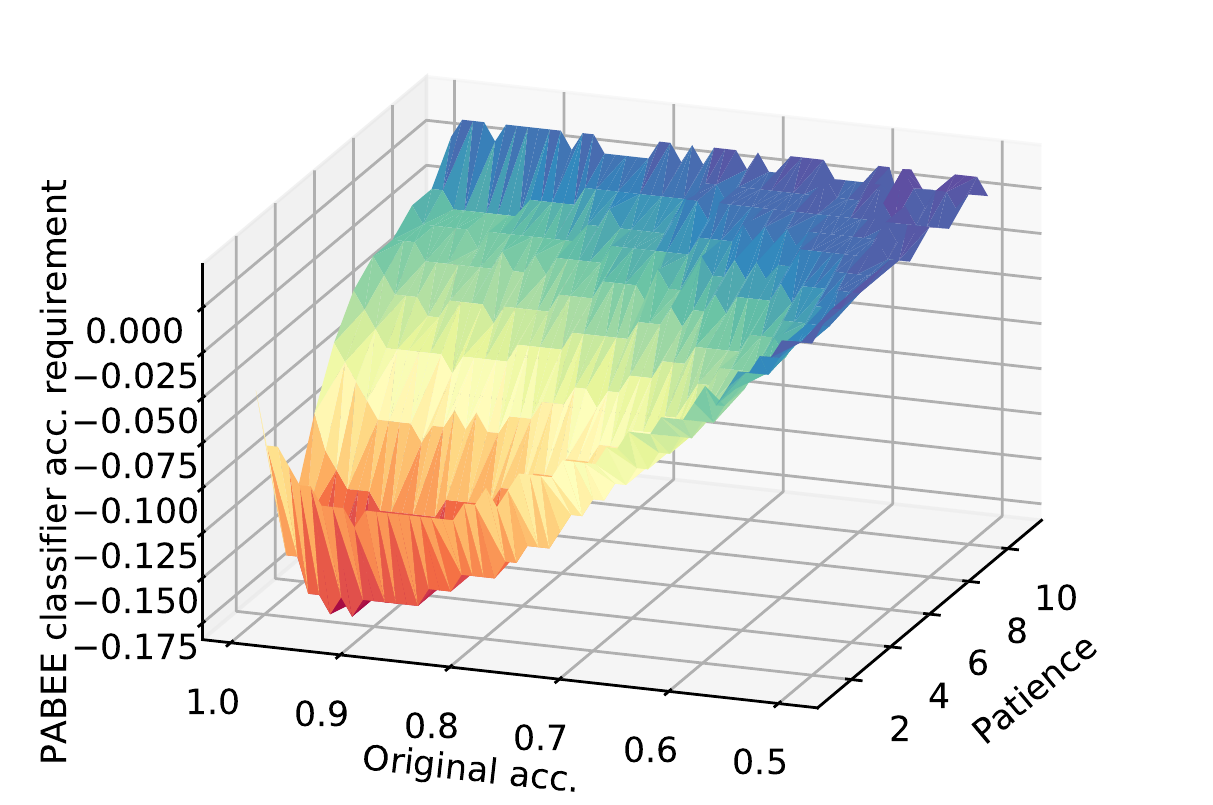}
        \caption{\label{fig:reducedacc}Accuracy requirement reduction effect of \abbr classifiers.}
    \end{subfigure}
    \caption{\label{fig:montecarlo}Monte Carlo simulation of per \abbr classifier's accuracy vs. the original inference accuracy under different patience settings.}
\end{figure}

Shown in Figure \ref{fig:acclb}, we illustrate the accuracy lower bound of each single $C_i$ needed for \abbr to reach the same accuracy as the original accuracy without \abbr. We run the simulation 10,000 times with random Bernoulli Distribution sampling for every $0.01$ of the original accuracy between $0.5$ and $1.0$ with patience $t\in[1,11]$. The result shows that \baby can effectively reduce the needed accuracy for each classifier. Additionally, we illustrate the accuracy requirement reduction in Figure \ref{fig:reducedacc}. We can see a valley along the patience axis which matches our observation in Section \ref{sec:patience}. However, the best patience in favor of accuracy in our simulation is around $3$ while in our experiments on real models and data suggest a patience setting of $6$. To analyze, in the simulation we assume all classifiers have the same accuracy while in reality the accuracy is monotonically increasing with more layers involved in the calculation.

\end{document}